\documentclass{article} 
\usepackage{iclr2026_conference,times}

\usepackage{amsmath,amsfonts,bm}

\def\eqref#1{equation~\ref{#1}}

\def\1{\bm{1}}

\DeclareMathAlphabet{\mathsfit}{\encodingdefault}{\sfdefault}{m}{sl}
\SetMathAlphabet{\mathsfit}{bold}{\encodingdefault}{\sfdefault}{bx}{n}

\usepackage{hyperref}
\usepackage{url}

\usepackage[utf8]{inputenc} 
\usepackage[T1]{fontenc}    
\usepackage{booktabs}       
\usepackage{amsfonts}       
\usepackage{nicefrac}       
\usepackage{microtype}      
\usepackage{xcolor}         

\usepackage{graphicx} 
\usepackage{array}      
\usepackage{hhline}     
\usepackage{mathrsfs}
\usepackage{caption}        
\usepackage{makecell}       
\usepackage[table]{xcolor}  
\usepackage{amsmath}
\usepackage{multirow}
\usepackage{tcolorbox}
\usepackage{colortbl}
\usepackage{amssymb}
\usepackage{bbding} 
\usepackage{pifont}         
\usepackage{nicematrix} 
\usepackage{cite}
\usepackage{enumitem}

\definecolor{lightgray}{RGB}{240,240,240}
\definecolor{lightblue}{RGB}{220,235,255}
\definecolor{Warmorange}{RGB}{255,222,189}
\definecolor{bluelevelone}{RGB}{240,250,255} 
\definecolor{blueleveltwo}{RGB}{227,239,255} 
\definecolor{bluelevelthree}{RGB}{215,232,255} 
\definecolor{mygray}{gray}{.9}

\title{Impostor: An Agent-Curated Benchmark for Realistic AIGC Manipulation Localization}

\author{Zhenliang Li$^{1,}$\thanks{Equal contribution.}\hspace{0.4em},
Yutao Hu$^{1,*,}$\thanks{Corresponding author.}\hspace{0.4em},
Qixiong Wang$^{2,*}$, Wenpeng Du$^{1}$, 
Hongxiang Jiang$^{2}$, \\
\textbf{Jiasong Wu$^{1}$}, 
\textbf{Xiaolong Jiang$^{2}$,}
\textbf{Jungong Han$^{3}$}
\\
$^{1}$Southeast University
\quad
$^{2}$Xiaohongshu Inc.  
\quad
$^{3}$Tsinghua University  
}

\iclrfinalcopy 
\begin{document}

\maketitle

\lhead{Preprint}

\vspace{-2.3mm}
\begin{abstract}
Recent advances in generative image editing have improved the realism and controllability of localized image manipulation, raising new challenges for image manipulation detection and localization (IMDL). However, existing IMDL benchmarks still have limitations in visual realism, manipulation diversity, and generator coverage, making it difficult to reflect recent trends in image manipulation. To address these limitations, we introduce Impostor, a high-quality AI-edited image manipulation localization dataset containing 100K manipulated images. Impostor is constructed by CraftAgent, a closed-loop agent framework that integrates scene perception, editing planning, manipulation execution, quality validation, and iterative reflection to automatically generate diverse and visually realistic manipulated images. Moreover, Impostor contains images generated by seven recent AIGC models across three manipulation types and includes multiple manipulated regions, providing a more comprehensive benchmark for AIGC-based IMDL. Furthermore, we propose PhaseAware-Net (PANet), a semantic-forensic framework that introduces local phase modeling and semantic-forensic consistency learning to better localize semantically plausible yet forensically disrupted manipulated regions. Extensive experiments show that Impostor poses significant challenges to existing large vision-language models (LVLMs) and specialized IMDL methods, while PANet achieves superior performance on Impostor and multiple public benchmarks.
\vspace{-2.5mm}
\end{abstract}

\section{Introduction}
\label{sec:intro}
\begin{figure*}[t]
    \centering
    \includegraphics[
        width=\textwidth,
        trim=-0mm 0mm 0mm 0mm,
        clip
    ]{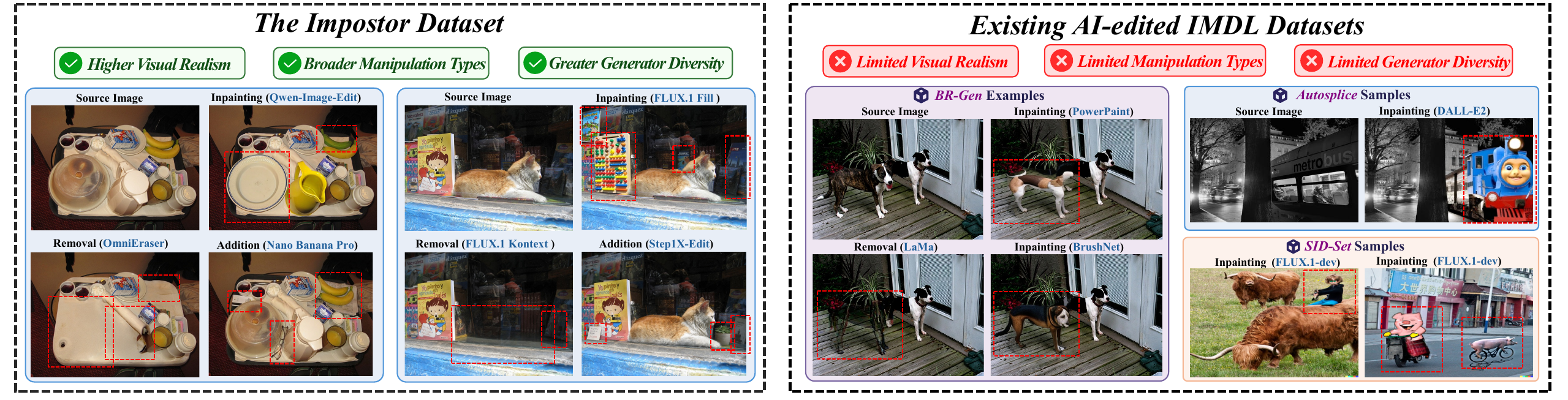}
    \captionsetup{
        font=small,
        labelfont=small,
        textfont=small
    }
    \vspace{-5mm}
    \caption{
    Comparison between Impostor and existing AIGC manipulation localization datasets. Impostor better addresses visual realism, manipulation diversity, and generator coverage, providing a more representative benchmark for modern AIGC-based image manipulations.
    }
    \label{fig:compare_dataset}
    \vspace{-5mm}
\end{figure*}

\vspace{-0.5mm}
Recent advances in generative image editing models \citep{labs2025flux,qwenimage} have made localized manipulations increasingly controllable and semantically consistent with the surrounding context, posing new challenges for image manipulation detection and localization (IMDL). However, existing IMDL benchmarks \citep{BRGAN,sidset} cannot fully reflect recent trends in image manipulation. Specifically, as shown in Figure \ref{fig:compare_dataset}, existing datasets face three key limitations: (i) \textbf{limited visual realism}, where samples are often generated with simple editing instructions and the manipulated region could easily be found; (ii) \textbf{limited manipulation diversity}, where datasets cover only a few manipulation types and mainly focus on single-region edits; and (iii) \textbf{limited generator coverage}, where datasets only include limited generators. Therefore, existing benchmarks cannot fully assess whether IMDL methods remain effective in modern AIGC-based manipulation scenarios.


Moreover, to achieve accurate IMDL, extracting high-quality forensic cues, such as residual noise and frequency statistics, plays an important role in prior works \citep{mvss,catnet}. However, existing works still face two limitations in AIGC-based manipulation scenarios. First, as suggested in studies \citep{chai2020makes, patch}, forensic cues in manipulated regions may be disrupted, leading to weaker consistency with semantic representations. In contrast, forensic cues extracted from authentic regions are generally more consistent with their semantic representations. However, existing IMDL methods often simply fuse semantic and forensic representations without explicitly supervising such consistency. Second, many IMDL methods rely on global frequency-domain analysis, which overlook local phase structures disrupted by AIGC editing and is less sensitive to the disruption of local forensic cues. Overall, these limitations weaken the model's ability to distinguish subtle and visually plausible manipulated regions generated by advanced AIGC models.

To address these limitations, on the one hand, we introduce Impostor, a high-quality benchmark comprising 100K manipulated images generated by recent AIGC models. Impostor is designed to better approximate realistic localized editing scenarios along three axes: \textbf{higher visual realism}, \textbf{greater manipulation diversity}, and \textbf{broader generator coverage}. For \textbf{visual realism}, Impostor is constructed by CraftAgent, a closed-loop agent framework that combines user-intent-aligned instruction generation with strict quality control. Specifically, instruction generation provides high-quality editing prompts to guide AIGC models in performing localized edits that better align with user editing intents in real-world scenarios. The quality control step further filters out samples with obvious visual artifacts and semantically implausible manipulations. In addition, all manipulated images undergo human Turing-style verification, where each manipulated image is required to be judged as real by at least two annotators, further ensuring the visual realism of the dataset from a human perception perspective. For \textbf{manipulation diversity}, Impostor covers three representative manipulation types and multi-region manipulation scenarios. For \textbf{generator coverage}, Impostor integrates seven state-of-the-art image editing models, covering diverse visual characteristics and manipulation traces from recent AIGC editing models. As shown in Table~\ref{tab:summary}, the significant degradation in LVLM detection performance on Impostor further supports the perceptual realism and deceptive nature of Impostor.


On the other hand, we propose PANet, a two-stream semantic-forensic framework for AIGC-based manipulation localization. Motivated by the above observation, PANet introduces a Forensic-Semantic Consistency (FSC) loss to explicitly supervise the consistency between semantic representations and forensic cues. FSC encourages authentic regions to produce stronger semantic-forensic consistency responses, enabling subtle forensic cues to better guide the localization of visually plausible yet forensically disrupted manipulated regions. To support this consistency learning with reliable forensic cues, we further design a Local Forensic Cue Extraction Module (LFCEM), which builds upon conventional residual cues and introduces local phase modeling to capture fine-grained structural forensic cues overlooked by global frequency-domain analysis. Finally, the semantic and forensic branches are integrated through a Mutual Guided Cross Attention Module (MGCAM) for multi-scale information interaction and accurate manipulation localization.

In summary, our main contributions are as follows:
\begin{itemize}[topsep=0pt, partopsep=0pt, itemsep=2.5pt, parsep=1pt,leftmargin=2.5em]

\item We construct \textbf{Impostor}, a high-quality AIGC manipulation localization dataset built by \textbf{CraftAgent}, with higher visual realism, greater manipulation diversity, and broader generator coverage.

\item We propose \textbf{PANet}, a two-stream semantic--forensic framework that combines local forensic cue extraction with explicit semantic-forensic consistency supervision, improving the localization and generalization capacity of subtle AIGC manipulations.

\item Extensive experiments show that \textbf{Impostor} poses greater challenges to both LVLMs and specialized IMDL methods, while \textbf{PANet} outperforms state-of-the-art methods in both localization and cross-dataset generalization.

\end{itemize}
\vspace{-0.3mm}

\section{Related Works}
\subsection{AI-generated Manipulation Localization Datasets}
Current IMDL benchmarks lag behind the rapid evolution of generative AI, failing to adequately reflect modern localized image manipulations generated by advanced  AIGC models. Early benchmarks such as CASIA \citep{casia}, Columbia \citep{Columbia}, and HiFi-IFDL \citep{hifi-net} focus on conventional manipulation types, such as splicing and copy-move, which tend to exhibit relatively fixed artifact patterns. While foundational for traditional IMDL research, they are not suited to modern generative editing scenarios. With the development of diffusion models (DMs)  \citep{ho2020denoising} and text-guided editing models \citep{2023instructpix2pix}, CocoGlide \citep{glideturfor} and AutoSplice \citep{autosplice} are among the earliest to introduce diffusion-based inpainting into the IMDL setting, but remain limited in scale and rely on single generators. The advent of large foundation models enables automated and scalable benchmark construction. GIM \citep{GIM} expands to the million level through a pipeline assisted by SAM \citep{sam2} and LLMs, but its editing scenarios are largely restricted to salient objects from ImageNet \citep{imagenet} and simple object-level substitutions. More recently, BR-Gen \citep{BRGAN} and SID-Set \citep{sidset} incorporate advanced DMs but similarly rely on coarse and rigid  category-replacement instructions that ignore scene context and user editing intent.  OpenSDI \citep{opensdi} adopts relatively more refined editing instructions, yet lacks closed-loop quality control. Moreover, all of these datasets mainly focus on single-region manipulation and cover limited manipulation types. As shown in Figure \ref{fig:compare_dataset}, none of these benchmarks fully addresses these three key limitations.

\vspace{-0.2mm}
\subsection{AI-generated Manipulation Localization Methods}
\vspace{-0.2mm}
In IMDL, existing methods typically formulate detection and localization as a joint task with dual-stream architectures. Early approaches \citep{mvss,pscc,catnet} rely heavily on priors tailored to conventional manipulations (e.g., JPEG inconsistency and boundary gradient anomalies), which are less effective for high-quality AIGC edits characterized with globally coherent reconstruction and naturally blended semantic boundaries. More recently, large pretrained models such as CLIP \citep{clip} have significantly improved localization performance, but their gains largely come from model-scale and pretraining advantages rather than explicit modeling of generative mechanisms. Several studies \citep{li2025learnable,nam2025m2sformer} attempt to explore manipulation traces left by generative processes from a frequency-domain perspective. However, they mainly focus on global statistical anomalies and remain limited for fine-grained localized manipulation analysis.  Meanwhile, they do not explicitly transform phase-related structural cues into local representations that are spatially aligned with manipulated regions. Moreover, existing methods often overlook the intrinsic relation between high-level semantic structures and low-level forensic cues in AIGC edits. These limitations motivate us to introduce spatially aligned local phase modeling and semantic-forensic consistency learning for fine-grained manipulation localization.

\vspace{-0.35mm}
\section{Impostor Dataset}
\begin{figure*}[t]
    \centering
    \includegraphics[
        width=0.98\textwidth,
    ]{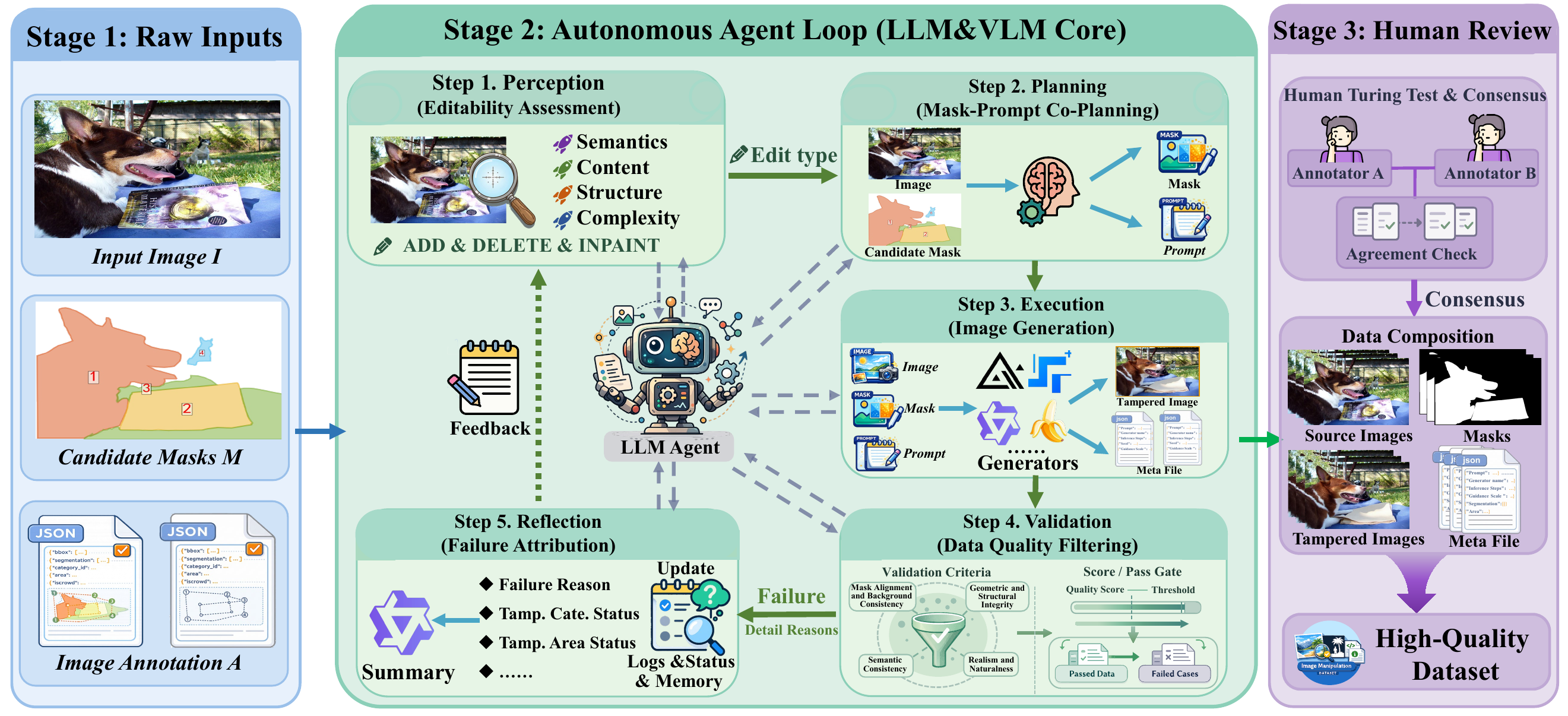}
    \vspace{-1mm}
    \caption{The generation pipeline of Impostor. An LVLM-based CraftAgent iteratively perceives, plans, executes, validates, and reflects to generate diverse and high-quality manipulated images.}
    \label{fig:dataset}
    \vspace{-5mm}
\end{figure*}

\vspace{-0.35mm}
Rather than simply scaling up data scale, Impostor is designed to address three key limitations of existing AIGC manipulation localization benchmarks: limited visual realism, limited manipulation diversity, and limited generator coverage. To address the heavy human effort and lack of automatic iterative refinement in conventional sequential construction pipelines, we construct Impostor using \textbf{CraftAgent}, a closed-loop agent framework designed to generate manipulated images that are both visually realistic and reliably annotated. The generation pipeline of Impostor is shown in Figure~\ref{fig:dataset}.

\vspace{-0.35mm}
\subsection{Source Dataset Selection}
\vspace{-0.35mm}
We select LVIS \citep{lvis} as the real image source, as its long-tail category coverage, complex natural scenes, and fine-grained mask annotations naturally support object-level manipulations across diverse semantic contexts and multiple regions. This choice also mitigates shortcut bias in existing IMDL datasets \citep{GIM,BRGAN} built from foreground-centric sources such as ImageNet, where models may localize salient objects rather than genuine manipulated regions.

\vspace{-0.35mm}
\subsection{CraftAgent: A Closed-Loop Data Construction Framework}
\vspace{-0.35mm}
Impostor is constructed using CraftAgent, a closed-loop agent framework designed to better simulate user editing behavior in practical scenarios. Inspired by the reasoning-and-acting paradigm of ReAct \citep{agent_react}, CraftAgent iteratively performs perception, planning, execution, validation, and reflection during dataset construction. Due to space limitations, we provide an overview of the agent procedure here. The detailed descriptions of each step are provided in the Appendix.
 

\noindent\textbf{User-intent-aligned instruction generation. \ }
Unlike existing AIGC manipulation benchmarks \citep{sidset,BRGAN}, which often generate manipulated images with simple instructions and limited scene understanding, perceptually realistic manipulation should better simulate how users edit images in practice. For an image, a user usually first decides the editing goal, then selects a suitable region, and finally describes the desired change according to the local content and surrounding context. Therefore, CraftAgent first performs a \textbf{Perception} step to analyze scene content and target-region properties, enabling the selection of a suitable manipulation type, such as addition, removal, or inpainting. It then performs a \textbf{Planning} step to generate a region-specific instruction describing the manipulation type and fine-grained visual attributes of the target object, such as shape, material, color, and texture. As suggested by prior work \citep{manas2024improving}, prompt quality strongly affects generation quality. Therefore, high-quality localized editing requires instructions that are both semantically plausible and well aligned with the target region, local structure, and scene context. To support this process, we employ advanced LVLMs \citep{OpenAI2025GPT51, bai2025qwen3} as the core decision-making module of CraftAgent. In addition, We introduce feedback-guided planning. At iteration $t$, the reflection module summarizes feedback $\mathcal{F}_t$ from previous iterations, including category distribution, manipulation type counts, and failure attribution, to guide the next planning step. This mechanism can reduce error accumulation and improve dataset diversity.

\noindent\textbf{Multi-Generator Manipulation Construction. \ }
Existing AIGC manipulation benchmarks are often built with limited editing models, restricting manipulation diversity and encouraging detectors to rely on generator-specific shortcuts. To address this issue, Impostor integrates seven state-of-the-art image editing models: six open-source models, including FLUX.1 Fill \citep{flux2024}, FLUX.1 Kontext \citep{labs2025flux}, PowerPaint \citep{powerpaint}, Step1X-Edit \citep{liu2025step1x}, Qwen-Image-Edit \citep{qwenimage}, and OmniEraser \citep{wei2025omnieraser}, and one closed-source model, Nano Banana Pro \citep{deepmind2025nanobananapro}. Given the manipulated region and editing instruction generated during the planning stage, CraftAgent performs an \textbf{Execution} step to invoke suitable editing models for manipulated image generation. This design introduces more diverse manipulation patterns into the dataset, enabling more comprehensive evaluation of localization performance for IMDL methods.

\noindent\textbf{Quality Validation and Reflection. \ }
Ensuring both visual realism and spatial alignment is central to AIGC manipulation localization dataset construction. A valid sample should appear realistic while keeping the edited content strictly aligned with the annotated mask. However, generative editing models may fail to follow the instruction or modify regions outside the target mask, compromising sample validity and making IMDL evaluation unreliable. Therefore, after image generation, CraftAgent performs a \textbf{Validation} step to assess each sample from four aspects: semantic consistency, mask alignment and background consistency, geometric and structural integrity, and overall realism. Samples are then analyzed through a \textbf{Reflection} step, where the agent summarizes failure attribution and other feedback signals to guide subsequent planning and generation. This closed-loop process reduces visually unrealistic samples and improves annotation reliability. Detailed validation criteria are provided in the Appendix.

{
\setlength{\aboverulesep}{0.2pt}%
\setlength{\belowrulesep}{0.2pt}%
\renewcommand{\arraystretch}{1.7}%
\newcommand{\rot}[1]{\rotatebox[origin=c]{70}{\scriptsize #1}}
\begin{table*}[t]
\scriptsize
\setlength{\tabcolsep}{2.1pt} 

\centering
\captionsetup{
    font=small,
    labelfont=small,
    textfont=small
}
\caption{
Summary of existing dataset attributes. Traditional manipulation types such as splicing and copy-move are categorized as ``Add''. ``Categories'' denotes the number of basic semantic categories of manipulated objects, with details provided in the Appendix. Right-side results are obtained through direct testing on manipulated images only, where * indicates the corresponding training source. ``w/o Agent'' denotes a sequential pipeline without quality control and feedback mechanisms. ``w/o HV'' removes only the final human verification step.
}

\label{tab:summary}
\vspace{-2mm}
\renewcommand{\arraystretch}{1.}
\begin{NiceTabular}{c c cc c c c ccc|cccc|cccc}[
]
    \Xhline{1.2pt}
    \rowcolor{lightgray}
         &  &  \multicolumn{2}{c}{Dataset Scale} &  &   &    & \multicolumn{3}{c|}{Manip. Type} &  \multicolumn{4}{c|}{Image-level Detection ACC} & \multicolumn{4}{c}{Pixel-level Localization IOU} 
         
    \\[-0.1ex]     
    \cmidrule[.4pt](lr){3-4}
    \cmidrule[.4pt](lr){8-10}
    \cmidrule[.4pt](lr){11-14}
    \cmidrule[.4pt](lr){15-18}
    
    \rowcolor{lightgray}  
   
        \multirow{-2}{*}{\rot{Type}} & \multirow{-2}{*}{\rot{DataSet}} & \rot{Real} & \rot{Fake} & \multirow{-2}{*}{\rot{Masks}} & \multirow{-2}{*}{\rot{Models}}  &  \multirow{-2.5}{*}{\rot{Categories}}  & \rot{Inpaint} & \rot{Add} & \rot{Remove} &  \rot{Qwen3VL} & \rot{FakeShield} & \rot{Gemini3Pro} & \rot{FatFormer} & \rot{FakeShield} & \rot{MVSS-Net} & \rot{NFA-ViT} & \rot{Sparse-ViT}\\ 
    \specialrule{.4pt}{0pt}{0pt} 

    \multirow{6}{*}{Manual}
      & Columbia &183 &180 &1 &- &- &\ding{55} &\ding{51}& \ding{55}
        & 95.00 &91.66 &97.71 &-   &69.58 &36.98 &33.59 &89.12 \\ 
        
      & CASIA V1  &800 &921 &1 &- &70 &\ding{55} &\ding{51}&\ding{55} &44.07& 92.01&64.24& -  &56.74&42.67&21.51&66.81 \\
        
      & CASIA V2 &7,491 &5,123 &1 &- &145 &\ding{55}& \ding{51}& \ding{55}
      &27.75&*75.82&31.61& -  & 48.38 &*66.72  &16.55&*70.80  \\
      
      & PSCC-Net  & 81,910 & 194,829 & 1   & - & 80 & \ding{55}& \ding{51}& \ding{51}   
      & 61.98& 98.19&59.15& -  &53.29&21.22&17.35&42.36  \\
      
      & SP COCO  & -   & 200,000 & 1 & - & 80 & \ding{55}& \ding{51}& \ding{55} 
      & 87.45& 71.08&90.87& -  &65.87&32.78&19.08&*83.51 \\

    & IMD2020  & 414    & 2,010   & 1-2   & - & 180 & \ding{55}& \ding{51}& \ding{55}    
    & 79.17&93.42& 83.75&- &    63.87 &22.59&19.04&*62.52\\
    
    \Xhline{0.8pt}
    \multirow{5}{*}{AIGC}
    
    & CocoGlide & -      & 512    & 1-2 & 1 & 73 & \ding{51}& \ding{55}& \ding{55} 
    & 67.18&100.00&51.95&44.15 &42.96 &22.56&64.21& 27.19\\
    
    & AutoSplice & 2,273  & 3,621  & 1 & 1 & 134 &  \ding{51}& \ding{55}& \ding{55} 
    & 44.05&57.73 &66.08&47.52 &48.36 &19.37&70.15& 25.14\\
    
    & SID-Set   & 100,000 & 100,000 & 1-2 & 1 & 300  &  \ding{51}& \ding{55}& \ding{55}  
    & 43.78&85.89 &50.33&11.58 &23.27 &16.75&39.43&17.90\\
    
    & BR-Gen  & 15,000 & 150,000    & 1 & 5 & 635 & \ding{51}& \ding{55}& \ding{51}  
    & 50.29&30.51 &65.49&42.68 &25.35 &9.85 &*90.71&11.13 \\ 
    \hhline{|~|-----------------|}
    
    \multirow{3}{*}{Impostor}
    \rowcolor{lightblue}
    & \textbf{w/o Agent}  & -  & -  & \textbf{1-9}  &  \textbf{7} & \textbf{-} & \ding{51}& \ding{51}& \ding{51}   
    &26.15&15.16&36.21& 9.12  &18.76 & 8.17 &21.15&8.25\\

    \rowcolor{lightblue}
    & \textbf{w/o HV}  & -  & -  & \textbf{1-9}  &  \textbf{7} & \textbf{-} & \ding{51}& \ding{51}& \ding{51}   
    &17.91&12.16&25.62& 7.79  &15.98 & 7.51 &19.51&6.91\\
    
    \rowcolor{lightblue}
    & \textbf{Full pipeline}  & 100,000  & 100,000  & \textbf{1-9}  &  \textbf{7} & \textbf{2800} & \ding{51}& \ding{51}& \ding{51}   &13.47&10.23   &19.85& 6.38 &14.23 &7.10&18.11&6.18 \\      
    \Xhline{1.2pt}
\end{NiceTabular}
\vspace{-2mm}
\end{table*}
}

\vspace{-0.6mm}
\subsection{Human Verification and Dataset Finalization}
\vspace{-0.6mm}
Since automated validation may still miss subtle issues in visual realism or annotation alignment, we further adopt human Turing-test-style verification as the final curation step. Each manipulated image is independently reviewed by two annotators through a two-stage protocol. In the first stage, annotators are shown only the manipulated image and judge whether it appears to be a naturally captured photograph. In the second stage, images judged as natural are further checked for mask alignment with the manipulated region. A sample is retained only when both annotators approve both its naturalness and mask alignment; otherwise, it is discarded. This final verification removes approximately 16\% of the automatically validated candidates, mainly due to visible artifacts or mask-region mismatch.

\vspace{-0.6mm}
\subsection{Benchmark Properties and Data Split} \label{sec:dataset_split}
\vspace{-0.6mm}
Impostor is a large-scale benchmark for AIGC manipulation localization, containing 100,000 samples, as summarized in Table \ref{tab:summary}. Each sample pair includes the manipulated image, source image, tampering mask, and generation metadata. Impostor has three key properties. First, higher visual realism: Impostor poses greater challenges to recent LVLMs, suggesting stronger perceptual realism and visual deception. Second, greater manipulation diversity: Impostor covers object addition, removal, and inpainting in multi-region manipulation scenarios, with up to nine manipulated regions per image. It also contains approximately 2,800 manipulated object categories, providing rich semantic diversity. Third, broader generator coverage: Impostor integrates seven advanced AIGC editing models, introducing diverse editing traces and visual distributions.


For dataset partitioning, samples from each generator are divided into training and test subsets with a 7:1 split, and each source image along with all associated manipulated images is assigned to the same subset to prevent data leakage. Moreover, the datasets used for comparison have different manipulated-to-real image ratios. Our experiments show that such ratio differences can significantly affect the cross-dataset generalization of IMDL models. To ensure fair comparison, we fix the training ratio of manipulated to real images at 6:1 across all training datasets.

\begin{figure*}[t]
    \vspace{-5mm}
    \label{model}
    \includegraphics[width=1\linewidth]{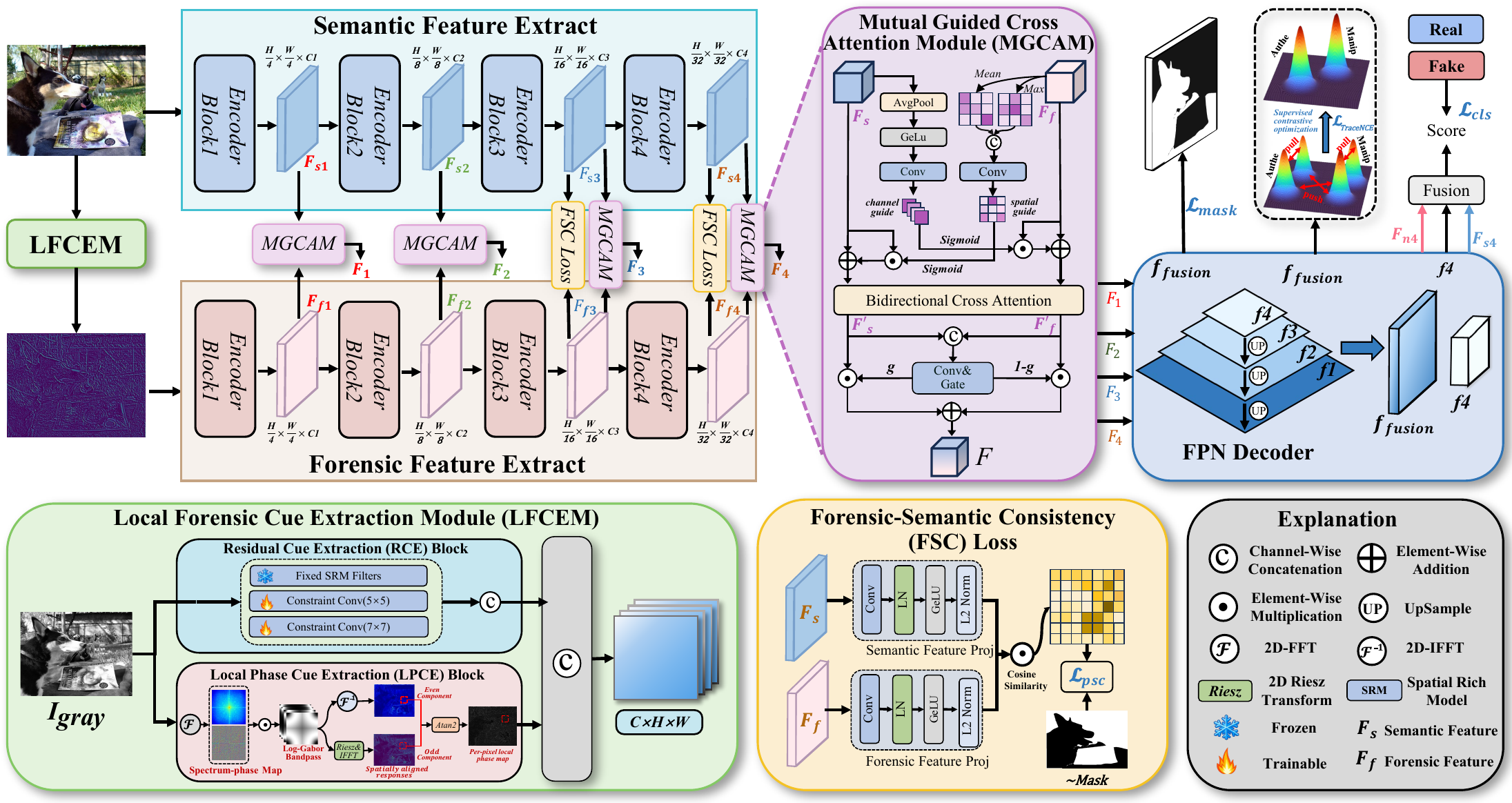}
    \vspace{-4mm}
    \caption{Overview of PANet. A semantic branch extracts multi-scale semantic features, while an LFCEM-enhanced forensic branch extracts residual and spatially aligned local phase cues. These features are fused via MGCAM and decoded by an FPN decoder. The FSC loss explicitly supervises semantic--forensic consistency, enabling subtle forensic cues to better guide manipulation localization.}
    \label{fig:model}
    \vspace{-2mm}
\end{figure*}

\vspace{-1mm}
\section{Methods}
\noindent\textbf{Overview. \ } 
Natural images exhibit structured low-level patterns, including texture, residuals, and frequency/phase statistics, which are correlated with object categories and scene structures ~\citep{simoncelli2001natural, torralba2003statistics}. In AIGC-based editing, the manipulated regions may remain semantically plausible, but the editing process can disturb low-level cues \citep{chai2020makes} and weaken their consistency with semantic representations. Motivated by these observations, we propose \textbf{PANet}, a semantic-forensic framework with two key designs: local forensic cue extraction and semantic-forensic consistency learning. Specifically, LFCEM extracts manipulation-sensitive residual and local phase cues, while FSC explicitly models the consistency between semantic and forensic representations. By encouraging authentic regions to exhibit stronger semantic-forensic consistency than manipulated regions, PANet enables subtle forensic discrepancies to guide accurate manipulation localization. Further diagnostic analyses in Appendix support the effectiveness of semantic-forensic consistency learning and local phase cue extraction.


\vspace{-0.5mm}
\subsection{Local Forensic Cue Extraction Module}
\vspace{-0.5mm}
To capture manipulation-sensitive local forensic cues, we introduce a Local Forensic Cue Extraction Module (LFCEM)  at the input of the forensic branch. LFCEM contains two parallel blocks: a Residual Cue Extraction (RCE) block that extracts residual-domain traces, and a Local Phase Cue Extraction (LPCE) block that captures phase-related local structural traces. Their outputs are concatenated into a hybrid forensic representation and then fed into the forensic encoder.

\vspace{-0.5mm}
\subsubsection{Residual Cue Extraction Block}
\vspace{-0.5mm}
RCE block extracts residual-domain forensic cues by combining fixed Spatial Rich Model (SRM) \citep{srm} filters with learnable constrained convolutions. Given an input image $I \in \mathbb{R}^{3 \times H \times W}$, we first apply frozen SRM filters as fixed high-pass priors to obtain residual responses. To capture task-adaptive residual patterns, we further introduce learnable constrained convolutions.  For a convolution kernel $W \in \mathbb{R}^{K \times K}$ with center position $(c,c)$, the constraint is defined as:
\begin{equation}
W_{c,c} = - \sum_{(u,v)\neq(c,c)} W_{u,v},
\end{equation}
which ensures a zero-sum kernel, $\sum_{u,v} W_{u,v}=0$, inducing a high-pass effect that suppresses constant or low-frequency content The final. residual representation is 
$F^{\mathrm{RCE}}=\operatorname{Concat}(F^{\mathrm{SRM}},F^{\mathrm{CC}})$.

\vspace{-0.5mm}
\subsubsection{Local Phase Cue Extraction Block}

\noindent\textbf{Motivation. \ }
RCE mainly captures high-frequency residual statistics, which reflect the strength of local responses but provide limited information about the spatial organization of local structures. Local phase complements residual cues by encoding structural coherence in edges, contours, and textures. Although realistic localized manipulations may remain subtle in residual statistics, they can still disrupt local phase coherence, motivating us to introduce spatially aligned local phase representations. To the best of our knowledge, LPCE is the first attempt to explicitly construct spatially aligned monogenic local phase cues for AIGC-based manipulation localization.

\noindent\textbf{Local Phase Cue Extraction \ } 
Given an RGB image, we denote its grayscale luminance map as $Y$ and compute $Y_f=\mathcal{F}(Y)$. Unlike global FFT/DCT-based methods, LPCE uses FFT only to obtain multi-scale band-limited responses and explicitly converts them into spatially aligned local phase maps through Riesz transformation and inverse Fourier transform. Specifically, we apply isotropic Log-Gabor band-pass filters $G_s$ at multiple scales $s$ to obtain scale-specific spectra:
\begin{equation}
B_s = Y_f \odot G_s,
\end{equation}
where \(\odot\) denotes element-wise multiplication. We then construct a 2D monogenic signal via the Riesz transform \citep{monogenic}. The even response and two odd responses at scale \(s\) are defined as:
\begin{equation}
f^{even}_s = \operatorname{Re}\big(\mathcal{F}^{-1}(B_s)\big), \quad
h^{odd}_{x,s} = \operatorname{Re}\big(\mathcal{F}^{-1}(B_s \odot \mathcal{R}_x)\big), \quad
h^{odd}_{y,s} = \operatorname{Re}\big(\mathcal{F}^{-1}(B_s \odot \mathcal{R}_y)\big),
\end{equation}
where \(\mathcal{R}_x\) and \(\mathcal{R}_y\) are frequency-domain Riesz multipliers along two orthogonal directions, defined as \(\mathcal{R}_x(u,v)=-i\,u/\sqrt{u^2+v^2+\epsilon}\) and \(\mathcal{R}_y(u,v)=-i\,v/\sqrt{u^2+v^2+\epsilon}\). Here, \(\operatorname{Re}(\cdot)\) denotes taking the real part. Note that after the inverse Fourier transform, the resulting feature maps are complex-valued and we take their real parts for subsequent computations. The local odd magnitude $h_s$, local amplitude $A_s$, and local phase $\phi_s$ are computed as:
\begin{equation}
h_s = \sqrt{{h^{odd}_{x,s}}^2 + {h^{odd}_{y,s}}^2 }, \qquad
A_s = \sqrt{{f^{even}_s}^2 + {h^{odd}_{x,s}}^2 + {h^{odd}_{y,s}}^2}, \qquad
\phi_s = \operatorname{atan2}(h_s, f^{even}_s).
\end{equation}
Here, \(A_s\) measures local structural response strength, while \(\phi_s\) encodes the local structural type at the current scale. Instead of using a single phase map, we construct a four-channel phase descriptor:
\begin{equation}
P_s = \operatorname{Concat}\!\left(\cos \phi_s,\ \sin \phi_s,\ \log(1 + A_s),\ \phi_s\right).
\end{equation}
Here, \(\cos \phi_s\) and \(\sin \phi_s\) provide a wrapping-robust phase encoding, \(\log(1+A_s)\) provides structural strength, and \(\phi_s\) retains the raw phase response. Finally, multi-scale phase descriptors are concatenated and aggregated by a lightweight convolutional module:
\begin{equation}
P = \operatorname{Concat}(P_1, P_2, \dots, P_S), \qquad
F^{\mathrm{LPCE}} = \operatorname{ConvGNGeLU}(P),
\end{equation}
In this way, LPCE constructs multi-scale phase-aware forensic representations, capturing local structural consistency cues that are difficult to reveal through residual magnitude alone.

\subsection{Mutual Guided Cross Attention Module}
To enhance semantic--forensic interaction, we design a Mutual Guided Cross Attention Module (MGCAM). Given the semantic feature \(F_s\) and forensic feature \(F_f\), MGCAM first generates a channel guide \(M_c\) from \(F_s\) and a spatial guide \(M_p\) from \(F_f\):
\begin{equation}
M_c =  \sigma\big(Conv\big(GeLu(\mathrm{GAP}(F_s)\big)\big),
\qquad
M_p = \sigma\big({Conv}\big([\mathrm{Avg}_c(F_f), \mathrm{Max}_c(F_f)]\big)\big),
\end{equation}
where \(\sigma(\cdot)\) denotes sigmoid activation; \(M_c\) encodes the importance of semantic-aware channel response, while \(M_p\) encodes the importance of forensic-sensitive spatial response.
\begin{equation}
\hat{F}_s = F_s + F_s \odot M_p, \qquad
\hat{F}_f = F_f + F_f \odot M_c,
\end{equation}
After mutual guidance, we apply bidirectional cross attention to allow each branch to attend to the other, further enhancing cross-branch feature interaction, which is expressed as follows:
\begin{equation}
F'_s = \mathrm{Attn}(Q_s, K_f, V_f), \qquad
F'_f = \mathrm{Attn}(Q_f, K_s, V_s),
\end{equation}
Finally, the two attended features are combined through an adaptive gating mechanism, which dynamically balances the contributions from each branch:
\begin{equation}
g = \sigma(\mathrm{Conv}_{1\times1}([F'_s,F'_f])),
\qquad
F = g \odot F'_s + (1-g) \odot F'_f,
\end{equation}
where \(g\) denotes the spatially adaptive gate. This adaptive fusion allows the model to dynamically balance semantic context and forensic cues for each spatial location.

\subsection{Loss Function}
The overall training objective consists of two primary supervision losses, $\mathcal{L}_{\mathrm{seg}}$ and $\mathcal{L}_{\mathrm{cls}}$, and two auxiliary losses, $\mathcal{L}_{\mathrm{fsc}}$ and $\mathcal{L}_{\mathrm{TraceNCE}}$. Specifically, $\mathcal{L}_{\mathrm{seg}}$ supervises localization via hybrid Dice-BCE loss, while $\mathcal{L}_{\mathrm{cls}}$ supervises classification via BCE loss. In addition, $\mathcal{L}_{\mathrm{fsc}}$ enforces Forensic-Semantic consistency, and $\mathcal{L}_{\mathrm{TraceNCE}}$ enhances region-level discriminability between authentic and manipulated features in the decoder space. Details are provided in the Appendix.

\section{Experiments}

\subsection{Experimental Setup}
\noindent\textbf{Datasets and Evaluation Protocols. \ }
To systematically evaluate both model generalization and the transferability of Impostor as a training source, we adopt two complementary protocols. Protocol I focuses on \textbf{cross-dataset pixel-level localization generalization}: models are trained on Impostor and directly tested on CoCoGlide \citep{glideturfor}, AutoSplice \citep{autosplice}, SID-Set \citep{sidset}, and BR-Gen \citep{BRGAN}. Protocol II evaluates \textbf{cross-domain image-level detection generalization} under the standard AIGCDetectBenchmark \citep{patch} setting, where models are trained on one source domain and tested on multiple unseen synthetic image domains. For Protocol I, all localization methods are trained and evaluated under the unified IMDL-BenCo framework \citep{imdlbench}  to ensure fair comparison. For Protocol II, we follow the official AIGCDetectBenchmark protocol and directly report the maintained benchmark records for compared image-level detection methods. For pixel-level localization, we report F1-score (F1) and Intersection over Union (IoU) on manipulated images only, since pixel-wise metrics are ill-defined for real images with zero-valued masks. For image-level detection, we report F1 and accuracy (Acc) on balanced sets containing equal numbers of real and manipulated images. Detailed descriptions of the compared methods and protocol settings are provided in the Appendix.


\noindent\textbf{Implementation Details. \ }
We use SegFormer-B2 \citep{segformer} as the semantic encoder, ConvNeXt V2-Tiny \citep{convnext} as the forensic encoder, and the Feature Pyramid Network (FPN) decoder for mask prediction. We train PANet for 45 epochs using the AdamW optimizer, with an input resolution of $512\times512$ and a total batch size of 45. The initial learning rate is set to $2\times10^{-4}$ with a weight decay of $5\times10^{-3}$, and the learning rate is scheduled with 3 warm-up epochs followed by cosine annealing. For data augmentation, we follow standard practice \citep{opensdi,BRGAN} and apply Gaussian blurring, JPEG compression, random scaling, and horizontal/vertical flipping. All training is conducted on two NVIDIA H100 GPUs.

\subsection{Comparison with State-of-the-art models}

\subsubsection{Cross-Dataset Pixel-level Localization Generalization}\label{sec:Cross-Dataset-Localization}
Table \ref{tab:crosss-dataset-segmentation} reports the Protocol I results on four external benchmarks and the Impostor test set, with the overall average calculated over all five datasets. PANet achieves the best average performance, obtaining an IoU/F1 of 0.5978/0.6603 and outperforming all comparison methods in overall cross-dataset localization. On the four external benchmarks, PANet achieves the best results on SID-Set \citep{sidset}, BR-Gen \citep{BRGAN}, CocoGlide \citep{glideturfor}, and AutoSplice \citep{autosplice}, with IoU/F1 scores of 0.4643/0.5338, 0.3269/0.3641, 0.6636/0.7319, and 0.7482/0.8198, respectively. On the in-domain Impostor test set, PANet also achieves the best IoU/F1 of 0.7859/0.8517. We attribute these gains to the collaborative modeling of semantic structures and forensic cues. The semantic branch provides scene-level and structural priors, while the forensic branch captures manipulation-sensitive residual and local phase traces. Their interaction, further guided by semantic-forensic consistency learning, enables PANet to better localize subtle abnormalities introduced by AIGC manipulations. Moreover, the strong cross-dataset performance of models trained on Impostor suggests that Impostor can serve as an effective and transferable training source for IMDL.

\subsubsection{Cross-Domain Image-Level Detection Generalization}\label{sec:Cross-Dataset-Detection}
Cross-domain image-level AIGC detection requires knowledge learned from a single source domain to generalize to unseen generation domains. Unlike pixel-level localization, this setting requires detectors to capture domain-robust forensic cues while avoiding overfitting to high-level semantic content in the source domain. Therefore, we adapt PANet to this protocol with task-specific modifications. Specifically, following recent AIGC detection studies~\citep{patch,Chameleon}, we divide each image into \(32\times32\) patches and select texture-rich and texture-poor patches as forensic inputs.  This patch selection only changes the forensic input construction, aiming to expose low-level artifacts that are more stable across generators.  For the semantic branch, we use a frozen CLIP ViT-Large~\citep{clip} encoder to provide domain-robust high-level representations, following the common practice in cross-domain AIGC detection.  The rest of the PANet framework remains unchanged. 

Under this setting, we evaluate the cross-domain image-level detection on AIGCDetectBenchmark \citep{patch}. As shown in Table \ref{tab:crosss-dataset-detection}, PANet achieves the best average performance, with a mean accuracy of 92.68, outperforming all comparison methods and showing a stronger generalization to unseen generation domains. Although PANet does not achieve the highest score on every individual dataset, it exhibits consistently strong performance across all datasets. Notably, most methods achieve their lowest or near-lowest accuracy on Impostor, highlighting its challenge even in image-level detection. Despite this difficulty, PANet achieves the highest accuracy on Impostor, reaching 68.91. This task-specific adaptation provides complementary evidence that PANet's core design remains effective beyond pixel-level localization.

{
\setlength{\aboverulesep}{0.15pt}%
\setlength{\belowrulesep}{0.15pt}%
\begin{table*}[t]
\caption{Pixel-level localization performance under Protocol I. \textbf{Bold} and \underline{underline} denote the best and second-best results, respectively.}
\label{tab:crosss-dataset-segmentation}
\scriptsize
\setlength{\tabcolsep}{3pt}   
\renewcommand{\arraystretch}{1.17}
\vspace{-1mm}
\centering
\begin{NiceTabular}{cc c >{\columncolor{lightblue}}c >{\columncolor{lightblue}}c  cc cc cc cc  cc}[
]

\Xhline{1.2pt}
\rowcolor{lightgray} &  &  & 
\multicolumn{2}{>{\columncolor{lightblue}}c}{Impostor} &
\multicolumn{2}{c}{SID-Set} &
\multicolumn{2}{c}{BR-Gen} & \multicolumn{2}{c}{CocoGlide}  &
\multicolumn{2}{c}{AutoSplice} & 
\multicolumn{2}{c}{Average}

    \\    
    \cmidrule[.4pt](lr){4-6}
    \cmidrule[.4pt](lr){6-7}
    \cmidrule[.4pt](lr){8-9}
    \cmidrule[.4pt](lr){10-11}
    \cmidrule[.4pt](lr){12-13}
    \cmidrule[.4pt](lr){14-15}
    
\rowcolor{lightgray}
\multirow{-2}{*}{Method}  & \multirow{-2}{*}{Year} &  \multirow{-2}{*}{Params} &  IoU & F1 & IoU & F1 & IoU & F1 & IoU & F1 & IoU & F1 & IoU & F1 \\

\specialrule{.4pt}{0pt}{0pt}
MVSS-Net   & 2021 & 146.9M   & 0.5000 & 0.5931 & 0.2789 & 0.3458 & 0.1297 & 0.1689 & 0.4240 & 0.5187 & 0.2938   & 0.3915 & 0.3253 & 0.4036 \\

CAT-NET  & 2022 &  114.3M  & 0.6342 & 0.6763 & 0.4181 & 0.4974 & 0.1893 & 0.2375 & 0.4722 & 0.5598 & 0.2861 & 0.3721 & 0.4000 & 0.4686 \\

ObjectFormer & 2022 & 130.6M   & 0.5419 & 0.6553  & 0.2482  & 0.3162 & \underline{0.3030} & \underline{0.3575} & 0.5235 & 0.6211   & 0.6814 & 0.7791 & 0.4596  & 0.5458 \\

TruFor  & 2023 & 68.7M  & 0.5117 & 0.5989 & 0.2231 & 0.2740 & 0.1685 & 0.3004 & 0.3621 & 0.5066 & 0.1669 & 0.2262 & 0.2865 & 0.3812 \\

IML-VIT  & 2024 & 91.8M  & 0.5799 & 0.6763 & \underline{0.4078} & \underline{0.4936} & 0.2563 & 0.3007 & 0.3440 & 0.4175 & 0.3601 & 0.4648  & 0.3896 & 0.4706 \\

Mesorch  & 2025 &  85.8M  & \underline{0.7079} & \underline{0.7876}  & 0.3908 & 0.4446 & 0.2457 & 0.2806 & \underline{0.6423} & \underline{0.7161} & 0.7030 & 0.7758 & \underline{0.5379} & \underline{0.6009} \\

M2SFormer  & 2025 & 28.1M  & 0.5350 & 0.6424 & 0.3397 & 0.4064 & 0.2438 & 0.3076 & 0.5323 & 0.6182 & 0.5441 & 0.6527 & 0.4390 & 0.5255 \\

NFA-ViT & 2026 & 28.2M    & 0.6751 & 0.7584 & 0.3150 & 0.3605 & 0.2441 & 0.2810 & 0.5326 & 0.6050 & \underline{0.7123} & \underline{0.7891} & 0.4958 & 0.5588 \\
\rowcolor{lightblue}
\textbf{PANet} & - & 74.9M  & \textbf{0.7859} & \textbf{0.8517} &\textbf{0.4643} & \textbf{0.5338} & \textbf{0.3269} & \textbf{0.3641} & \textbf{0.6636} & \textbf{0.7319} & \textbf{0.7482} & \textbf{0.8198}  & \textbf{0.5978} & \textbf{0.6603} \\
\Xhline{1.2pt}
\end{NiceTabular}
\end{table*}
}

\newcommand{\rot}[1]{\rotatebox[origin=c]{70}{\scriptsize #1}}
\begin{table*}[t]
\vspace{-1mm}
\scriptsize
\caption{Cross-domain image-level detection results under Protocol II on AIGCDetectBenchmark\citep{patch}.  \textbf{Bold} and \underline{underline} denote the best and second-best results, respectively.}
\vspace{-2mm}
\label{tab:crosss-dataset-detection}
\setlength{\tabcolsep}{1.8pt}   
\renewcommand{\arraystretch}{1.17}
\centering
\makebox[\textwidth][c]{
\begin{tabular}{c cccccccccccccccc >{\columncolor{lightblue}}c c}
\Xhline{1.2pt}
\rowcolor{lightgray}
Method & \rot{ProGAN} & \rot{StyleGAN} & \rot{BigGAN} & \rot{CycleGAN} &
\rot{StarGAN} & \rot{GauGAN} & \rot{StyleGAN2} & \rot{WFIR} &
\rot{ADM} & \rot{Glide} & \rot{Midjourney} &
\rot{SD v1.4} & \rot{SD v1.5} & \rot{VQDM} & \rot{Wukong} & \rot{DALLE2} &
\cellcolor{lightblue}\rot{Impostor} & \rot{Mean} \\
\specialrule{.4pt}{0pt}{0pt}

CNNSpot    & 100.00 & 90.17 & 71.17 & 87.62 & 94.60 & 81.42 & 86.91 & 91.65 & 60.39 & 58.07 & 51.39 & 50.57 & 50.53 & 56.46 & 51.03 & 50.45 & 50.27 & 69.57 \\
FreDect    & 99.36  & 78.02 & 81.97 & 78.77 & 94.62 & 80.57 & 66.19 & 50.75 & 63.42 & 54.13 & 45.87 & 38.79 & 39.21 & 77.80 & 40.30 & 34.70 & 33.85 &62.25 \\
Fusing     & \textbf{100.00} & 85.20 & 77.40 & 87.00 & 97.00 & 77.00 & 83.30 & 66.80 & 49.00 & 57.20 & 52.20 & 51.00 & 51.40 & 55.10 & 51.70 & 52.80 & 49.77 &67.29 \\
Gram-Net & 99.99 & 87.05 & 67.33 & 86.07 & 95.05 & 69.35 & 87.28 & 86.80 & 58.61 & 54.50 & 50.02 & 51.70 & 52.16 & 52.86 & 50.76 & 49.25 & 48.12 &  69.64 \\

LNP       & 99.95 & 92.64 & 88.43 & 79.07 & \textbf{100.00} & 79.17 & 93.82 & 50.00 & 83.91 & 83.50 & 69.55 & 89.33 & 88.81 & 85.03 & 86.39 & 92.45 & 55.74 & 83.39 \\
UnivFD     & 99.81  & 84.93 & \underline{95.08} & \underline{98.33} & 95.75 & \textbf{99.47} & 74.96 & 86.90 & 66.87 & 62.46 & 56.13 & 63.66 & 63.49 & 85.31 & 70.93 & 50.75 & 50.78 &76.80 \\
DIRE-G     & 95.19  & 83.03 & 70.12 & 74.19 & 95.47 & 67.79 & 75.31 & 58.05 & 75.78 & 71.75 & 58.01 & 49.74 & 49.83 & 53.68 & 54.46 & 66.48 & 50.94 &68.68 \\
PatchCraft & \textbf{100.00} & 92.77 & \textbf{95.80} & 70.17 & \underline{99.97} & 71.58 & 89.55 & 85.80 & 82.17 & 83.79 & \textbf{90.12} & \textbf{95.38} & \textbf{95.30} & 88.91 & 91.07 & \textbf{96.60} & 59.27 &89.31 \\
NPR        & 99.79  & 97.70 & 84.35 & 96.10 & 99.35 & 82.50 & \textbf{98.38} & 65.80 & 69.69 & 78.36 & 77.85 & 78.63 & 78.89 & 78.13 & 76.11 & 64.90 &  53.18 & 81.16 \\
AIDE & \underline{99.99} & \textbf{99.64} & 83.95 & \textbf{98.48} & 99.91 & 73.25 & 
\underline{98.00} & \underline{94.20} & \underline{93.43} & \textbf{95.09} & 77.20 & 93.00 & 92.85 & \textbf{95.16} & \underline{93.55} & \textbf{96.60} & 67.01 & \underline{91.25}\\ 


\rowcolor{lightblue}
PANet & \underline{99.99} & \underline{99.19} & 90.61 & 96.33 & 99.77 & \underline{86.43} & \textbf{98.38} & \textbf{97.50} & \textbf{93.79} & \underline{94.29} & \underline{81.22} & \underline{93.56} & \underline{93.29} & \underline{93.97} & \textbf{93.79} & \underline{94.60} & \textbf{68.91} & \textbf{92.68} \\

\Xhline{1.2pt}
\end{tabular}
}
\end{table*}

\subsection{Robustness and Ablation Studies of PANet}
\vspace{-1mm}
To simulate common post-processing degradations during image transmission and storage, we evaluate the robustness of models under various distortions. Specifically, we apply resizing to different scales, Gaussian blur with varying kernel sizes, Gaussian noise with different standard deviations, and JPEG compression with varying quality factors. As shown in Table \ref{tab:robustness_experiments}, PANet performs the most robustly across different distortions, validating its potential for real-world applications.

Table \ref{ablation_phasenet} reports the ablation study of PANet. The full model achieves the best average IoU/F1 of 0.5978/0.6603. For forensic representation ablation, replacing LFCEM with raw RGB input causes a clear drop of 3.74 IoU points, confirming the necessity of explicit forensic cue extraction. More importantly, LPCE-only and RCE-only outperform the raw-RGB baseline by 2.13 and 2.30 IoU points, respectively, while the full LFCEM further improves over both. This demonstrates that local phase and residual cues are both effective and complementary forensic representations. For framework-level components, removing the forensic branch causes the greatest degradation, indicating that semantic features alone are insufficient for robust localization. Removing MGCAM also degrades performance, showing the benefit of explicit semantic-forensic interaction. Notably, removing FSC decreases IoU from 0.5978 to 0.5840, indicating that explicit semantic-forensic consistency supervision provides meaningful additional guidance for accurate manipulation localization. Finally, TraceNCE loss brings a smaller but consistent improvement by  enhancing authentic--manipulated feature separability.

\begin{table*}[t]
\begin{center} 
\begin{minipage}[t]{0.5\linewidth} 
    \vspace{0pt}
    \scriptsize
    \setlength{\tabcolsep}{2pt}
    \renewcommand{\arraystretch}{1.15}
    \centering
        \captionsetup{
        font=small,
        labelfont=small,
        textfont=small
    }
    \caption{Robustness analysis under common post-processing distortions, measured by Average IoU. }
    \vspace{-3mm}
    \label{tab:robustness_experiments}
    \begin{NiceTabular}{ccccc}[
    ]
    \Xhline{1.2pt}
        
    \rowcolor{lightgray}
    Distortion & NFA-ViT & M2SFormer & Mesorch & PANet\\
    \specialrule{.8pt}{0pt}{0pt}
    no distortion     & 0.4870 & 0.4390 & \underline{0.5379} & \textbf{0.5978} \\
    Resize(0.78×)     & 0.4435 & 0.4185 & \underline{0.5311} & \textbf{0.5825}  \\
    Resize(0.25×)     & 0.3152 & 0.3348 & \underline{0.3633} & \textbf{0.4505} \\
    Blur($k=3$)  & 0.4375 & 0.4491 & 
    \underline{0.5050} & \textbf{0.5756} \\
    Blur($k=7$)  & 0.3610 & \underline{0.3985} & 0.3933 & \textbf{0.4928} \\
    Noise($\sigma$ = 3) & 0.2057 & 0.3036& \underline{0.3907} & \textbf{0.4300}\\
    Noise($\sigma$ = 5)  & 0.1234 & 0.2187 & \underline{0.3277} & \textbf{0.3877}\\
    Compress($q$ = 95)    & 0.2828 & 0.3818 & \underline{0.4072} & \textbf{0.4878}\\
    Compress($q$ = 85)    & 0.2012 & \underline{0.3457} & 0.3071 & \textbf{0.3834}\\
    \Xhline{1.2pt}
    \end{NiceTabular}
    \vspace{-3mm}
\end{minipage}
\hfill  
\begin{minipage}[t]{0.48\linewidth}
    \vspace{0pt}
    \scriptsize
        \captionsetup{
        font=small,
        labelfont=small,
        textfont=small
    }
    \caption{Ablation study of PANet. }
    \vspace{-3mm}
    \label{ablation_phasenet}
    \setlength{\tabcolsep}{8pt}
    \renewcommand{\arraystretch}{1.15}
    \centering
    
    \begin{tabular}{ccc}
    \Xhline{1.2pt}
    
    \rowcolor{lightgray}
    \textbf{Variant} & \textbf{Average IoU} & \textbf{Average F1} \\
    
    \specialrule{.8pt}{0pt}{0pt}
    \rowcolor{lightgray}
    \multicolumn{3}{c}{\textbf{Level I: Forensic Representation Ablation}} \\
    
    w/o RCE (LPCE-only)   & 0.5817 & 0.6395 \\
    w/o LPCE (RCE-only)   & 0.5834 & 0.6404 \\
    w/o LFCEM (Raw RGB)  & 0.5604 & 0.6160 \\
    
    \specialrule{.8pt}{0pt}{0pt}
    \rowcolor{lightgray}
    \multicolumn{3}{c}{\textbf{Level II: Framework Component Ablation}} \\
    
    w/o Forensic Branch & 0.5443 & 0.6004 \\
    w/o MGCAM & 0.5741 & 0.6359 \\
    w/o FSC Loss       & 0.5840 & 0.6402 \\
    w/o TraceNCE Loss  & 0.5920 &  0.6491\\
    
    \specialrule{.8pt}{0pt}{0pt}
    \rowcolor{lightblue}
    \textbf{PANet} & \textbf{0.5978} & \textbf{0.6603} \\
    
    \Xhline{1.2pt}
    \end{tabular}
    \vspace{-4mm}

\end{minipage}
\vspace{-2mm}
\end{center}
\end{table*}

\vspace{-0.5mm}
\section{Conclusion}
\vspace{-0.5mm}
In this paper, we present \textbf{Impostor}, a high-quality benchmark for realistic AIGC-based image manipulation detection and localization (IMDL). By addressing the limited realism, manipulation diversity, and generator coverage of existing benchmarks, Impostor provides a more visually realistic and diverse benchmark that better reflects local editing behaviors in the real world. To further address the challenges posed by realistic localized manipulations, we further propose \textbf{PANet}, a two-stream framework that combines local phase modeling with semantic-forensic consistency learning for accurate manipulation localization. Extensive experiments show that Impostor poses substantial challenges to both LVLMs and specialized IMDL methods, while PANet achieves strong localization, robustness, and cross-dataset generalization. We hope that this work can serve as a useful benchmark and methodological foundation for future research on IMDL. 


\newpage
\bibliography{iclr2026_conference}
\bibliographystyle{iclr2026_conference}


\end{document}